\documentclass[review]{elsarticle}

\usepackage[dvipsnames]{xcolor}
\definecolor{LightGray}{rgb}{0.97,0.97,0.97}
\usepackage{listings} % syntax highlighting
\lstdefinelanguage{SPARQL}{
  basicstyle=\small\ttfamily,
  backgroundcolor=\color{LightGray},
  columns=fullflexible,
  breaklines=true,
  sensitive=true,
  frame=single,
  aboveskip=1em,
  belowskip=1em,
  xleftmargin=.5em,
  xrightmargin=.5em,
  framexleftmargin=.5em,
  framextopmargin=.5em,
  framexbottommargin=.5em,
  framexrightmargin=.5em,
  tabsize = 2,
  showstringspaces=false,
  morecomment=[l][\color{gray}]{\#},       % comments
  morecomment=[n][\color{blue}]{<http}{>}, % uris
  morestring=[b][\color{OliveGreen}]{\"},  % strings
  keywordsprefix=?,
  classoffset=0,
  keywordstyle=\color{Sepia},
  morekeywords={},
  classoffset=1,
  keywordstyle=\color{Purple},
  morekeywords={seas,rdf,rdfs,owl,xsd,purl, qudt, prov, sosa},
  classoffset=2,
  keywordstyle=\color{MidnightBlue},
  morekeywords={
    SELECT,CONSTRUCT,DESCRIBE,ASK,WHERE,FROM,NAMED,PREFIX,BASE,OPTIONAL,
    FILTER,GRAPH,LIMIT,OFFSET,SERVICE,UNION,EXISTS,NOT,BINDINGS,MINUS,a
  }
}

\usepackage[english]{babel}
\usepackage[T1]{fontenc}
\usepackage{etoolbox}
\makeatletter
\patchcmd{\ps@pprintTitle}{\footnotesize\itshape
      Preprint submitted to \ifx\@journal\@empty Elsevier
      \else\@journal\fi\hfill\today}{\scriptsize{Preprint submitted to Computers & Geosciences \hfill \today}}{}{}

\def\ps@pprintTitle{%
   \let\@oddhead\@empty
   \let\@evenhead\@empty
   \let\@oddfoot\@empty
   \let\@evenfoot\@oddfoot
}
\makeatother
\usepackage{comment}

\usepackage{subfig}
\usepackage{hyperref}
\usepackage{blindtext}
\usepackage{tcolorbox}
\usepackage{amsmath}

\usepackage{amsmath,amssymb,amsfonts}
\usepackage{algorithmic}
\usepackage[]{graphicx}
\graphicspath{{Figures/}}
\usepackage{lipsum}
\usepackage{balance}
\usepackage{textcomp}
\usepackage{multirow}
\usepackage{color}
\usepackage{lscape}
\usepackage{graphicx}
\usepackage{enumitem}
\usepackage{booktabs,siunitx,rotating}
\newcommand\atsign{@}
\NewDocumentCommand{\rot}{O{90} O{1em} m}{\makebox[#2][l]{\rotatebox{#1}{#3}}}%

\usepackage[switch]{lineno}

\usepackage{array}

\usepackage{nomencl}
\makenomenclature

\begin{document}

\begin{frontmatter}

\title{A semantic web approach to uplift decentralized household energy data}

\author[add1,add2]{Jiantao Wu}
\ead{jiantao.wu@ucdconnect.ie}
\author[add1,add3]{Fabrizio Orlandi}
\ead{fabrizio.orlandi@adaptcentre.ie}
\author[add4]{Tarek~AlSkaif}
\ead{tarek.alskaif@wur.nl}
\author[add1,add3]{Declan O'Sullivan}
\ead{declan.osullivan@adaptcentre.ie}
\author[add1,add2]{Soumyabrata Dev\corref{mycorrespondingauthor}}
\ead{soumyabrata.dev@ucd.ie}
\cortext[mycorrespondingauthor]{Corresponding author. Tel.: + 353 1896 1797.}

\address[add1]{The ADAPT SFI Research Centre, Ireland}
\address[add2]{School of Computer Science, University College Dublin, Ireland}
\address[add3]{School of Computer Science and Statistics, Trinity College Dublin, Ireland}
\address[add4]{Information Technology Group, Wageningen University and Research, The Netherlands}

\begin{abstract}
In a decentralized household energy system comprised of various devices such as home appliances, electric vehicles, and solar panels, end-users are able to dig deeper into the system's details and further achieve energy sustainability if they are presented with data on the electric energy consumption and production at the granularity of the device. However, many databases in this field are siloed from other domains, including solely information pertaining to energy. This may result in the loss of information (\textit{e.g.} weather) on each device's energy use. Meanwhile, a large number of these datasets have been extensively used in computational modeling techniques such as machine learning models. While such computational approaches achieve great accuracy and performance by concentrating only on a local view of datasets, model reliability cannot be guaranteed since such models are very vulnerable to data input fluctuations when information omission is taken into account. This article tackles the data isolation issue in the field of smart energy systems by examining Semantic Web methods on top of a household energy system. We offer an ontology-based approach for managing decentralized data at the device-level resolution in a system. As a consequence, the scope of the data associated with each device may easily be expanded in an interoperable manner throughout the Web, and additional information, such as weather, can be obtained from the Web, provided that the data is organized according to W3C standards.
\end{abstract}


\end{frontmatter}

\mbox{}
\nomenclature{\(HECP\)}{Household Energy Consumption and Production}
\nomenclature{\(NOAA\)}{National Oceanic and Atmospheric Administration}
\nomenclature{\(CoSSMic\)}{Collaborating Smart Solar-Powered Microgrids}
\nomenclature{\(ICT\)}{Information and Communications Technology}
\nomenclature{\(CA\)}{Climate Analysis}
\nomenclature{\(RDF\)}{Resource Description Framework}
\nomenclature{\(UTC\)}{Coordinated Universal Time}
\nomenclature{\(PV\)}{Photovoltaics}
\nomenclature{\(KG\)}{Knowledge Graph}
\nomenclature{\(PCC\)}{Pearson Correlation Coefficients}
\nomenclature{\(IoT\)}{Internet of Things}
\nomenclature{\(RDF\)}{Resource Description Framework}
\nomenclature{\(SAREF4EE\)}{the EEbus/Energy\atsign home extension of the Smart Appliances REFerence ontology}
\nomenclature{\(EKG\)}{Energy Knowledge Graph}
\nomenclature{\(SWRL\)}{Semantic Web Rule Language}
\nomenclature{\(GHG \)}{Greenhouse Gas}
\nomenclature{\(DR\)}{Demand Response}
\nomenclature{\(TMAX\)}{Maximum Temperature}
\nomenclature{\(SPARQL\)}{SPARQL Query Language for RDF}

\printnomenclature[1in]

\section{Introduction}
In light of the climate urgency with which humanity has been confronted in recent years, the European Commission 2021 Work Programme has established a ``Fit for 55'' package intended to reduce greenhouse gas (GHG) emissions by at least 55\% by 2030 and achieve a climate-neutral Europe by 2050~\cite{noauthor_undated-ri}. Among a variety of other considerations, energy efficiency is a major focus for the Union's ultimate decarbonization. This makes high energy efficiency a critical priority for all energy sectors, particularly the residential sector~\cite{jain2022deep}, which occupies more than a quarter of the Union's total final energy consumption. Energy decentralization has emerged as one of the most popular contemporary research topic in this domain as a mean for increasing energy efficiency~\cite{wu2021organizing}. With the growing usage of Information and Communication Technologies (ICT) in the Internet of Things (IoT) sector, data on household energy consumption and production (HECP) may now be generated in a decentralized manner, for example, from an electric vehicle, a heat pump, or home appliances. Due to the range and granularity of data-generating devices, a new generation of smart household energy systems is geared toward decentralization and has the potential to considerably assist in the transition to a sustainable energy future~\cite{van2020integrated,Xu2020-xg}. 

On the other hand, evaluating household energy data is getting increasingly difficult as a result of various smart devices interacting and forming a complex energy flow data network~\cite{jain2021validating,jain2020clustering}. Decentralized energy systems are often paired with research into data-driven technologies (\textit{e.g.} machine learning) for optimizing the systems based on the massive ocean of incoming data in order to manage the inherent risk associated with energy usage's intermittent and unpredictable nature and achieve energy sustainability, including cost reduction, emission reduction, and energy efficiency. However, most of those technologies are developed for project-specific decentralized data (\textit{i.e.} data is produced by a specific project) to solve problems of a specific energy sub-domain. A significant downside is that these technologies would fail to produce highly realistic and reliable results, as the energy sub-domain is interdependent with other domains~\cite{Teixeira2020-pb}. For example, solar energy generation is sensitive to weather condition \cite{alskaif2020systematic,Mussard2017-xb,dev2019estimating,dev2016estimation}. 

A key factor accounting for the data constraints in terms of cross-domain impact is the poor interoperability between the energy systems and other systems such as weather forecasts~\cite{orlandi2019interlinking, wu2021interoperable}. Assembling data of various sources and establishing interoperability among different heterogeneous systems can take much efforts, since data pieces published by different projects varies largely in terms of naming conventions, data formats(\textit{e.g.} CSV,JSON...), meta data, and \textit{etc}~\cite{wu2021detecting,wu2022rdfstr}. With the leading progress of energy studies in the last decade, many energy systems have been developed towards interoperability. This interoperability, however, is restricted to the transmission of messages across systems. As for interoperability at the knowledge level with respect to the domain, the data and features that may be made accessible and shared, which is currently advanced in many fields in connection to social networks and encyclopedias, is seldom studied for semantic enrichment of energy data~\cite{wu2021ontologyd}. Consequently, researchers are likely to get restricted data from a single project or to gather data within the confines of their limited efforts and then organize them for the purpose of building data-driven technologies\cite{ahmad2020review,wu2022boosting}. 
% This may result in a significant bias in the solution of small-scale smart energy issues, and therefore the model's reusability is harmed by a predicted more complex situation in which greater quantities, more kinds of data (e.g. weather data), and the trend toward decentralization in the smart energy space should be addressed as the network develops~\cite{ahmad2020review}.

Recently, semantic web researchers have made significant progress in establishing knowledge-level interoperability across data from diverse areas, such as climate analysis~\cite{wu2021automated,wu2022augment}. Relevant to their studies, ontology is critical in the semantic web since it serves as a specialized language for modeling domains shared by heterogeneous entities~\cite{Salatino2018-xm}. The incorporation of semantics into the data transmitted between parties enables a clear conception of the knowledge shared by both parties, hence increasing the effectiveness of data sharing by eliminating misunderstandings~\cite{wu2021uplifting,wu2022linkclimate}. Furthermore, the usage of semantic models results in other benefits, such as computational inference and knowledge reuse\cite{Hooda2020-zq}. They may be used to design systems that are not dependent on the data model, with a high degree of abstraction and flexibility that facilitates system growth, as well as to test the system's knowledge~\cite{wu2022vkg} and apply rules using, for example, the Semantic Web Rule Language (SWRL)~\cite{Horrocks2004-jw}.

In this work\footnote{\label{note3}In the spirit of reproducible research, the source code is available at \url{https://github.com/futaoo/semantic-energy}.}, we propose to address these problems by building strong ontology models and recognizing how cross-domain variables, such as weather data, affect energy consumption data. By using these technologies, it is feasible to conduct strategic explorations of semantic relationships between individual energy devices in a decentralized energy network, allowing knowledge to spread throughout the Web and be readily recognized by end users. Simultaneously, it enables the development of decentralized energy data with the capability of integrating any external data sources (e.g., a case study on meteorological data is presented in this work) without the need to re-model the data by redefining the schema or extending the fields of the table if the data is tabular in format. This will be further explored for the potential improvements on the reliability of today's data-driven~\cite{manandhar2019data,manandhar2018data} technologies in the smart energy field.
% We show how it is possible to integrate these heterogeneous data sources, usually distributed separately on the Web, facilitating data collection, fusion, and pre-processing. We demonstrate how a well-defined schema could be leveraged for integrating these datasets and supporting HEC research purposes.

In summary, this work's novel contributions include:

\begin{itemize}
    \item Creating systematic semantics for decentralized household energy consumoption and production data so as to grant the data knowledge-level interoperability;
    \item Converting household energy data to Linked Data~\cite{bizer2011linked} to simplify the integration of Web-wide semantified data from other domains;
    \item A case study to show that connecting external meteorological factors with energy consumption/production improves the data understanding and analysis.
\end{itemize}

The remainder of this article is structured as follows: Section~\ref{sec:relatedwork} gives the relevant literature that informs our study, including comparisons between our materials and those of others, as well as adoptions of other people's ideas. Section~\ref{sec:bkgd} contains material usage, including a description of the raw data and a number of key semantic methods utilized in this work. Section~\ref{sec:workflow} concentrates on the complete proposed workflow for transforming local household consumption data into Linked Data, which provides web-wide access plus knowledge-level interoperability to the local data. The benefits of Linked Data are shown via the augmentation of National Oceanic and Atmospheric Administration (NOAA) meteorological data with household energy data. Section~\ref{sec:analysishc} illustrates the use of local household energy data in a Linked Data platform by analyzing solar energy production in relation to temperature. Finally, in Section~\ref{sec:Conc}, we summarize our work and outline potential future endeavors.

\section{Related work}
\label{sec:relatedwork}
An emerging research perspective is to represent the data in an ontology model (defined in Section~\ref{sec:ontoshort}) such that the underlying relationships between data can be articulated in human words, bringing intelligence to data analysis. 

Earlier literature have developed application-specific ontologies and semantics-based systems. Abid et al.~\cite{Abid2018-mh} repurposed existing ontologies to develop a defect detection system capable of publishing user complaints of city problems such as water leaks and broken street lights as linked data to aid in the administration of smart cities. Synapse~\cite{An2020-kc} is a semantic web-based annotation system designed to enhance the metadata for all data in smart cities. Some famous applications such as Demand Response (DR) in residential energy systems are also benefited from many semantic enrichment studies~\cite{Cimmino2020-oj,Fernandez-Izquierdo2020-cf}. The unifying feature of these methods is that the generated models are oriented on the handling of real data, such as user input. The purpose of these models is to improve the introduction of diverse data sources in order to increase the system's completeness. However, many projects have released data for their own reasons; for example, in our work, the original goal of collecting CoSSMic (Collaborating Smart Solar-Powered Microgrids) energy statistics (for more information, see Section~\ref{sec:cossmic}), which are now historical records, was to investigate the smart grid system in a city. The fundamental issue addressed by this study is whether semantic web technologies may be utilized to harvest data that has been published for various reasons and then utilised to supplement our own research. This viewpoint is distinct from the majority of semantic technologies built on real data.

RDF (Resource Description Framework) is one of the most dependable models for combining different types of data in order to develop applications in the area of smart energy~\cite{Baken2020-ye}. The RDF models are often used in conjunction with Linked Data principles~\cite{bizer2011linked}, which are critical for data interoperability. Numerous academics have developed RDF-based semantics for data collected from smart energy systems. Chun et al.~\cite{Chun2020-su} designed Energy Knowledge 
Graph (EKG) to incorporate existing information about decentralized grids. Wagner~\cite{Wagner2010-xp} developed semantics for the privacy concerns of smart grid device usage. These models' major weakness is their incapacity to account for the effects of other domains. The interoperability of semantic technologies benefits just the smart energy sector. In contrast to these methods, our research will utilize an RDF knowledge model to integrate multi-domain data into fixed energy data.

Several researchers have previously published several general ontologies for energy data. The primary advantage of reusing a generic ontology is that it may be modified to serve particular purposes and therefore improve interoperability in semantic processing of the dataset. SAREF4EE (the EEbus/Energy\atsign home extension of the Smart Appliances REFerence ontology) is an ontology developed by Daniele at al.~\cite{noauthor_undated-gh} for the optimization of energy demand and response. SEAS was introduced by Lefran\c{c}ois~\cite{Lefrancois2017-ag}, with the goal of enabling interoperability across smart energy sectors. Our study is inspired by prior ontologies for smart energy systems, but it focuses on the decentralized household energy systems as well as ease of inclusion of cross-domain impacts (\textit{e.g.} climate domain), which has received less attention from other researchers. To complete the climatic impacts modeling portion of our model, we also referred to Wu's~\cite{wu2021ontology} ontology CA and Janowicz's~\cite{Janowicz2019-hn} ontology SOSA in order to describe the climatic sensor data.

\section{Overview of used data and technologies}
\label{sec:bkgd}
In this section, we deliver the several common sources of smart energy and climate data, as well as related semantic web technologies, as the preparation for the workflow description in Section~\ref{sec:workflow}. The energy data input, on the other hand, may be changed at whim to accommodate more broad research objectives in the context of smart energy development.
\subsection{Data description}
\label{sec:datadesc}
The next two sub-level sections provide a description of the experimental data utilized in this work's studies.
\subsubsection{CoSSMic household energy data}
\label{sec:cossmic}

CoSSMic\footnote{\url{http://isc-konstanz.de/en/isc/institute/public-projects/completed-projects/eu/cossmic.html}} is a smart grid project financed by the EU Framework Programme FP7 for Research and Innovation. Its objective is to optimize energy consumption in households in a German city—Konstanz—by creating intelligent microgrids~\cite{amato2017simulation}. The grid network's energy flow is managed by an autonomous ICT system that adapts energy consumption and distributed energy production in real time based on a variety of factors such as availability, pricing, and weather conditions. This decentralized energy flow is enabled through coordinated load shifting, in which power users and producers may negotiate an optimal energy exchange. The investigation's energy data is accessible on the open power system data site\footnote{\label{ftn:opsd}\url{https://data.open-power-system-data.org/household_data/}} and may be utilized for reanalysis. However, the current dataset only includes data on energy flow, and many other critical variables for understanding the energy exchange profile, such as weather data, are missing. The absence of these variables in our study will obstruct the reanalysis process. We try to address this issue in part in this work by republishing the data on the Web using semantic methods in order to acquire more data sources for reference. This approach will improve the usefulness of a dataset on local energy consumption.

\subsubsection{Link-climate knowledge graph}
\label{sec:linkclimate}

\begin{figure}[ht]
\centering
\includegraphics[clip, trim=0cm 7.5cm 0cm 4cm, width=\textwidth]{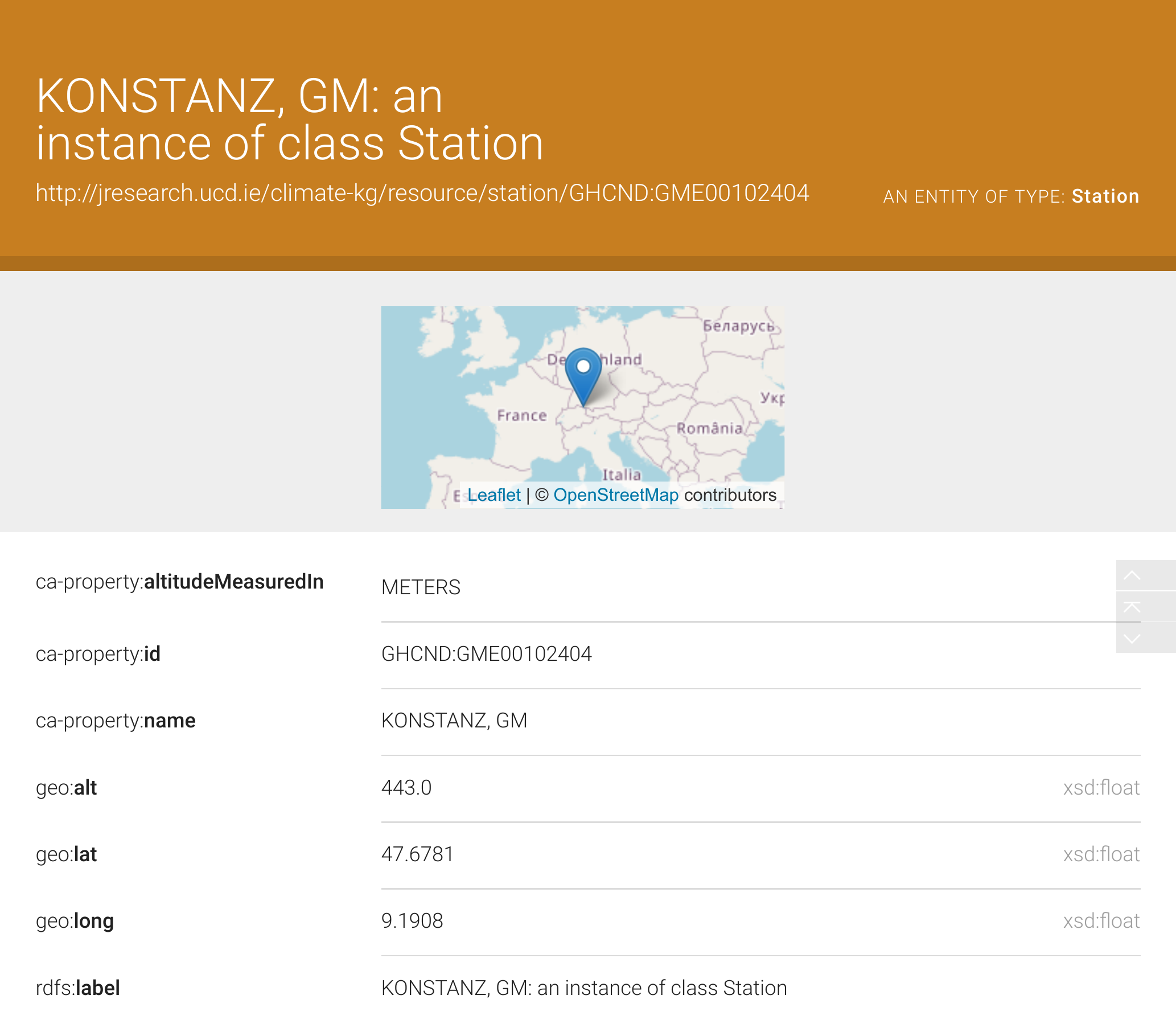} 
\caption{A Linked Data profile of a climate monitoring station in Konstanz}
\label{fig:konstation}
\end{figure}

Link-climate\footnote{\url{http://jresearch.ucd.ie/linkclimate/}} is a climate observation knowledge graph (KG) that adheres to the concepts of Linked Data (details are provided in Section~\ref{sec:shortsum})~\cite{wu2021ontology}. It offers NOAA Climate Online Data\footnote{\url{https://www.ncdc.noaa.gov/cdo-web/}} recorded by stations located in many European nations and cities (including Konstanz) through a Linked Data portal. Fig.~\ref{fig:konstation} illustrates a Linked Data representation of a climate station in Konstanz. The climate observation station's data set contains a variety of meteorological measurements, including temperature and precipitation. These data are accessible through the web and provide a flexible interface to any published Linked Data in any domain provided a suitable ontology model can make the connection.

\subsection{Useful semantic web approaches summarized}
\label{sec:shortsum}
This section will summarize the necessary semantic web technologies employed for the modeling purpose.

\subsubsection{Resource Description Framework}

RDF data model~\cite{Manola2004-jx} can be represented in a graph using: 1) a node for the subject, 2) an arc that goes from a subject to an object for the predicate and 3) a node for the object. In addition, uniform resource identifier (URI) is normally used to denote the subject, predicate, and object (SPO) in a RDF statement (except the blanked nodes). Currently, ontologies are usually coupled with the RDF data model to make semantic data interchange on the Web .

\subsubsection{Ontologies in short}
\label{sec:ontoshort}

According to the W3C definition\footnote{\url{https://www.w3.org/standards/semanticweb/ontology}}, ontologies (or vocabularies) clearly define words and connections for specific areas of interest. Take a wireless sensor network as an example, ontology can be used to simply refer to different kinds of sensors as ``precipitation sensor'', ``solar radiation sensor'', and so on. Similarly, relevant meteorological observation results may be connected to associated sensors through connections such as ``hasResults'' using the RDF SPO (subject, predicte, object) grammar. By specifying a set of words, a semantic layer of operational human-definable terms is constructed over the data.
% According to W3C's definition\footnote{\url{https://www.w3.org/standards/semanticweb/ontology}}: `ontologies' (or `vocabularies' in the simple term) unambiguously define concepts and relationships for specific fields of concern. Take wireless sensor network as an example, ontology can be made to classify different types of sensors by simply naming `wind sensor', `temperature sensor', etc. And the corresponding observations can be connected to by creating links named, for example, `hasResults'. Consequently with a set of terms defined, a semantic layer is formed consisting of operative human describable terms over the data. 
\subsubsection{Linked Data in short}
\label{sec:linkeddatashort}

Linked Data is an area of research that establishes a set of Linked Data principles for the organization, linking, and publication of content on the Web~\cite{bizer2011linked}: 1) use URIs as names for things; 2) use HTTP URIs so that people can look up those names; 3) when someone looks up a URI, provide useful information, using the RDF standard; and 4) include links to other URIs. It enables users to explore data by following data links between data sources, enabling a set collection of data sources to be easily extended with all of the other Linked Data sources available on the Internet~\cite{barbosa2021use}. In the last decade, Linked Data has been widely intertwined with ontologies to provide semantic interoperability across heterogeneous power systems. The increased number of inter-linked consistent power data sources has further boosted the inter-system data mining potential.  For instance, Gomes \textit{et al}.~\cite{Gomes2016-ju} published micro grid data as Linked Data to enable the examination of new services and algorithms for the administration of micro grids with the inclusion of interoperable data from various grid systems. Similarly, Wicaksono \textit{et al.}~\cite{Wicaksono2021-fg} incorporated multivariate DR (demand-response) data leveraging Linked Data and ontologies technologies to enhance the prediction results for machine learning applications. Another noticeable advantage of Linked Data principles is that federated SPARQL (SPARQL Query Language for RDF) queries can be performed on such a huge global database on the Web in order to acquire resources that are not available in fixed databases~\cite{8706177}. This can be achieved with SPARQL 1.1, which will be detailed in Section~\ref{sec:fedsparql}.

\section{Proposed workflow of linked decentralized energy data}
\label{sec:workflow}
The proposed workflow is to combine the CoSSMic decentralized energy data and link-climate data and then publish them as linked data. The ultimate goal of this process is to broaden the scope of a fixed CoSSMic data such that it can be queried together with the globe linked data on the web. We will begin by providing an overview of the process, followed by sub-level sections that explain each of the workflow's segmented components.
\subsection{Workflow overview}

\begin{figure}[ht]
\centering
\includegraphics[width=\textwidth]{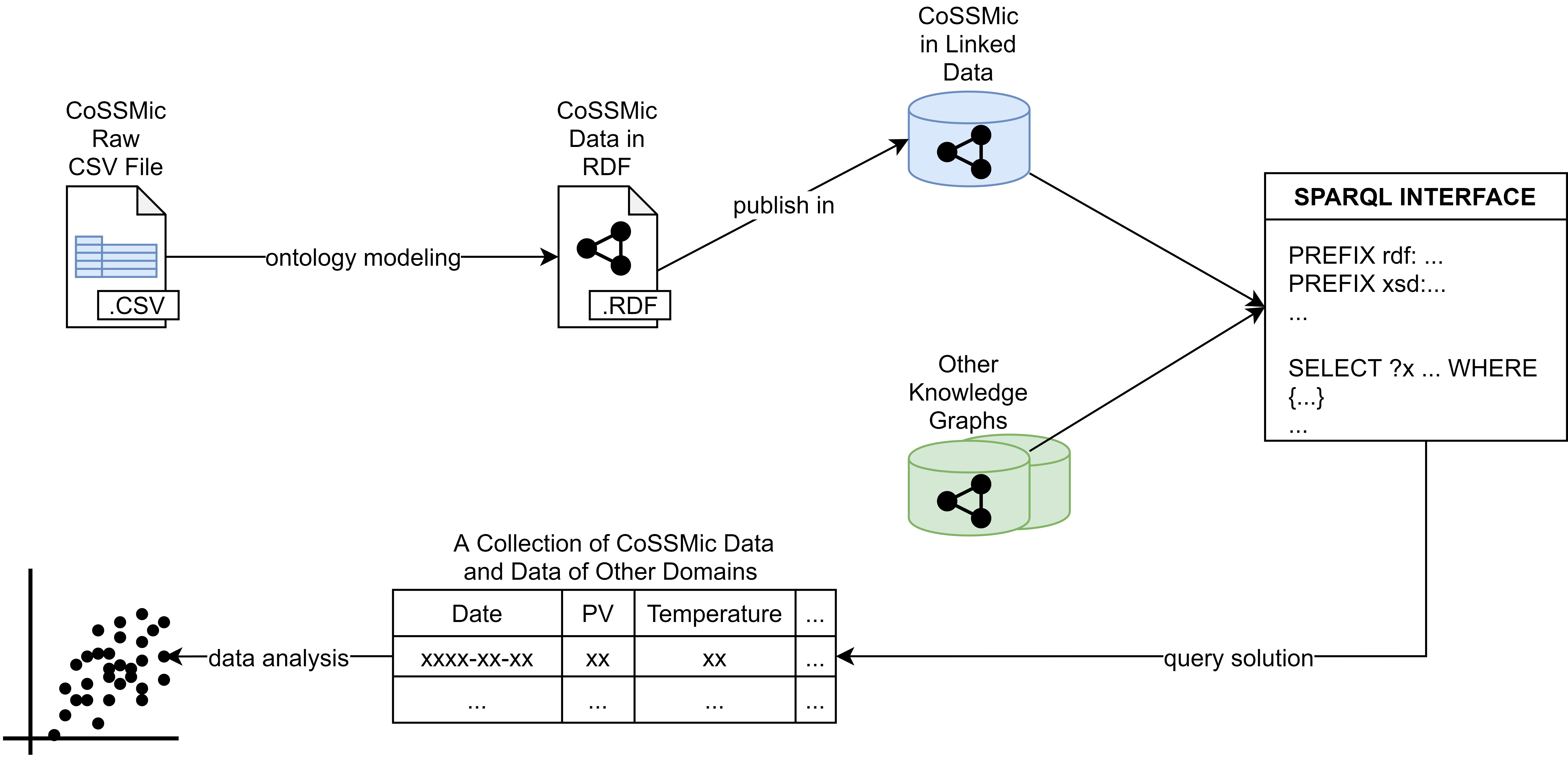} 
\caption{A schematic depiction of the whole workflow}
\label{fig:graph-workflow}
\end{figure}

At the beginning of this section, we provide an overview of the proposed process for semantic improvement of web-wide heterogeneous data (see Fig.~\ref{fig:graph-workflow} for a graphical representation). The ultimate goal is to enable Linked Data-based interoperability across different data sets, so that individual pieces of data may be easily joined for cross-domain analysis purposes leveraging Linked Data principles. The workflow's starting point is a dataset in any format (in our case, the chosen CoSSMic energy data is CSV formatted). Then, using ontology modeling, we construct semantics for the data, resulting in a KG in RDF. To be globally accessible through the web, the KG should adhere to Linked Data principles, such as naming nodes using URIs. Once the KG is published on the Internet, users can use SPARQL to query the CoSSMic KG in conjunction with KGs from other domains in order to augment the CoSSMic energy data for cross-domain analysis. The following sub-sections describe the workflow's components.

\subsection{Ontology modeling}
\label{sec: onto}
The ontology modeling process for the tabular CoSSMic datasets mainly consists of modeling the table headers (\textit{i.e.} data fields) and the data entries.
\subsubsection{Ontology modeling for table headings}

The open power system data set contains raw tabular CoSSMic data, as well as comprehensive description for the table headers. The purpose of this work is to first clarify the meaning of each column heading in the documentation and then to identify the connections between individuals that can be deduced from the table headers. For example, the heading ``DE\_KN\_industrial1\_pv\_1'' has the annotation ``Total photovoltaic energy generation in an industrial warehouse building in kWh''. A country—Germany (``DE''), a city—Konstanz (``KN''), an industrial building—industrial1 (``industrial1''), and a photovoltaic device—pv1 (``pv\_1'') can all be extracted. When each energy sector is seen as a system inside the energy network, the heading can therefore be rephrased as the semantic assertion that ``pv\_1'' is a subsystem of ``industrial1'', which is a industrial building in the German city of Konstanz. Following this approach, the set of table headers can be extended to include connections vertically between the individuals representations within each heading and also horizontally between individuals represented by separate headings. We utilize the SEAS knowledge model~\cite{seasd22} to characterize the possible individuals and their connections in order to create a complete ontology model for the CoSSMic data. The following are some of the most often used vocabulary for defining classes and attributes\footnote{\textbf{Note}: ontology vocabularies in this paper are already associated with web addresses and comply with the form \{prefix\}:\{literal term\} where the the meaning of the prefix (name space) is given in the supplementary graphical representations of the vocabularies.}.

\begin{figure}[ht]
\centering
\includegraphics[width=0.9\textwidth]{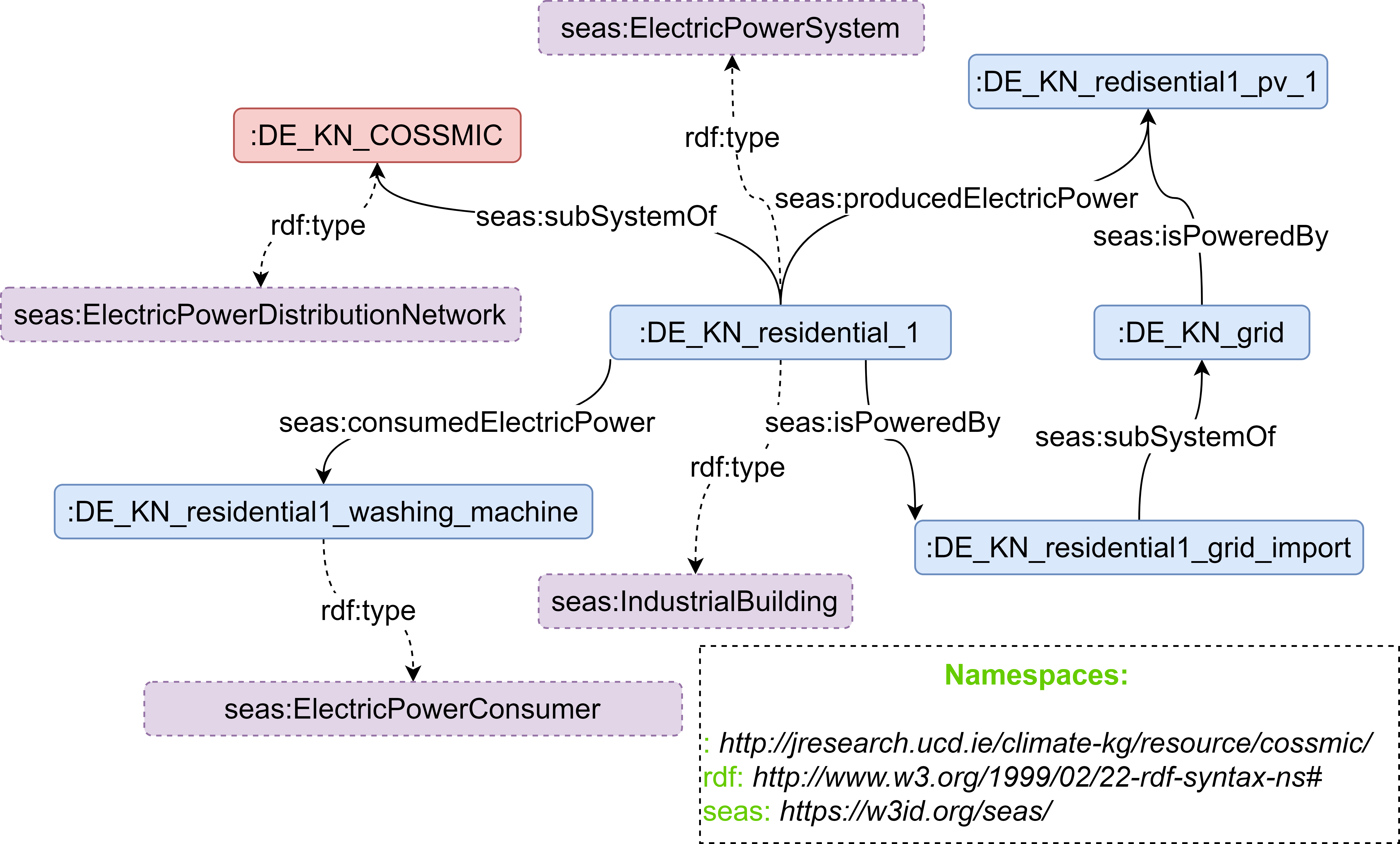} 
\caption{A graphical depiction of the ontology model for certain CoSSMic table headers, in which blue nodes represent the headings and purple nodes indicate classes; the whole CoSSMic electric power distribution network--``DE\_KN\_COSSMIC'' is colored as the red node.}
\label{fig:ontohead}
\end{figure}

\begin{itemize}
    \item \textbf{seas:ElectricPowerDistributionNetwork}~~ (CLASS) denotes a network used to distribute the electric power;
    \item \textbf{seas:ElectricPowerTransmissionSystem}~~ (CLASS) denotes an electric power transmission system capable of transmitting electricity;
    \item \textbf{seas:isPoweredBy}~~ (PROPERTY) links a System to its powered system and the inverse vocabulary is ``seas:powers'';
    \item \textbf{seas:producedElectricPower}~~ (PROPERTY) denotes the produced electric power;
    \item \textbf{seas:consumedElectricPower}~~ (PROPERTY) denotes the consumed electric power;
    \item \textbf{seas:subSystemOf}~~ (PROPERTY) links a system to its super system.
\end{itemize}
Listing~\ref{lst:rdf} illustrates the use of the aforementioned vocabulary to model (encoded in \texttt{Turtle} RDF format~\cite{beckett2014rdf}) some of the individuals retrieved from the table headers. The graphical representation of Listing~\ref{lst:rdf} is given by Fig.~\ref{fig:ontohead}.

\begin{lstlisting}[language=SPARQL,caption={From the public SPARQL endpoint, this example query gets a time series of CoSSMic HECP data with temperature alignments.},label=lst:rdf]
@base <http://jresearch.ucd.ie/climate-kg/ca/resource/cossmic/> .
@prefix rdf: <http://www.w3.org/1999/02/22-rdf-syntax-ns#> .
@prefix seas: <https://w3id.org/seas/> .

:DE_KN_residential_1
		rdf:type seas:IndustrialBuilding, seas:ElectricPowerSystem;
		seas:subSystemOf :DE_KN_COSSMIC;
		seas:producedElectricPower :DE_KN_residential1_pv;
		seas:consumedElectricPower :DE_KN_residential1_washing_machine;
		seas:isPoweredBy :DE_KN_residential_grid_import.
:DE_KN_COSSMIC
		rdf:type seas:ElectricPowerDistributionNetwork.
:DE_KN_residential_washing_machine
		rdf:type seas:ElectricPowerConsumer.
:DE_KN_residentia1_grid_import
		seas:subSystemOf :DE_KN_grid.
:DE_KN_grid
		seas:isPoweredBy :DE_KN_residential1_pv.
\end{lstlisting}

\subsubsection{Ontology modeling for data entries}
To convert the tabular data to RDF data for further storage as Linked Data, this phase continues to utilize the SEAS ontology to describe actual data objects, their connections, and their associated headings (data fields) in the CoSSMic dataset. This paper treats each record of energy consumption and production from each device as a ``Evaluation'' specified by SEAS ontology and makes use of the vocabularies ``seas:consumedElectricPower'' and ``seas:producedElectricPower'' to differentiate between energy consumption and production. The following are the primary vocabulary (see Figure 4 for a graphical illustration) for modeling data entries:
\begin{itemize}
    \item \textbf{seas:ElectricPowerEvaluation}~~ (CLASS) denotes evaluations for electric power properties;
    \item \textbf{seas:evaluation}~~ (PROPERTY) links a valuable entity to one of its evaluations.;
    \item \textbf{seas:evaluatedValue}~~ (PROPERTY) links an evaluation to the literal (numeric value in this paper);
\end{itemize}

\begin{figure}[ht]
\centering
\includegraphics[width=0.9\textwidth]{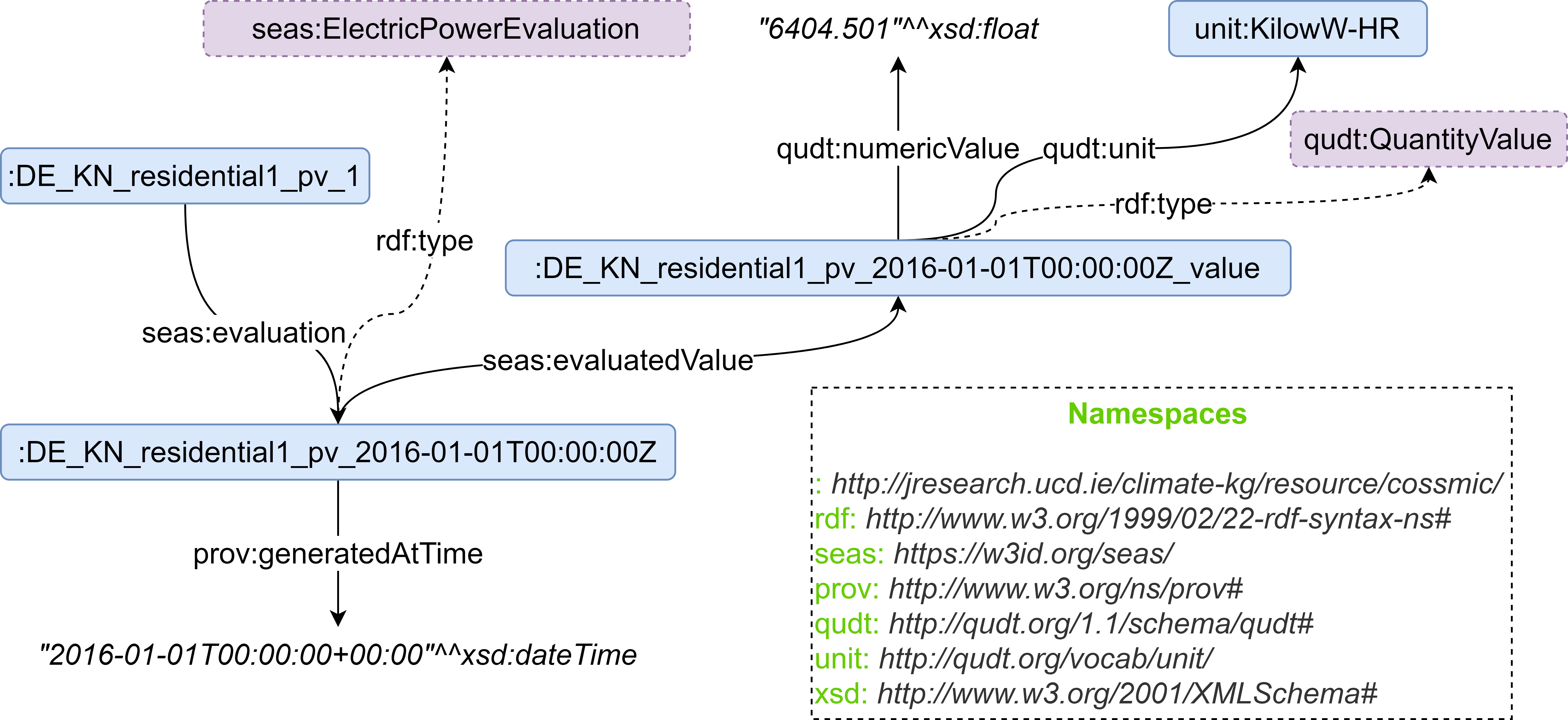} 
\caption{The node ``DE\_KN\_residential1\_pv'' has the identical real world entity representation as the node with the same name in Fig.~\ref{fig:ontohead}, and blue nodes are the record entities obtained from CoSSMic dataset; Coordinated Universal Time (UTC) is utilized for time.}
\label{fig:ontodata}
\end{figure}

\subsection{Uplifting NOAA data}
\label{sec:NOAA}

\begin{figure}[ht]
\centering
\includegraphics[width=0.9\textwidth]{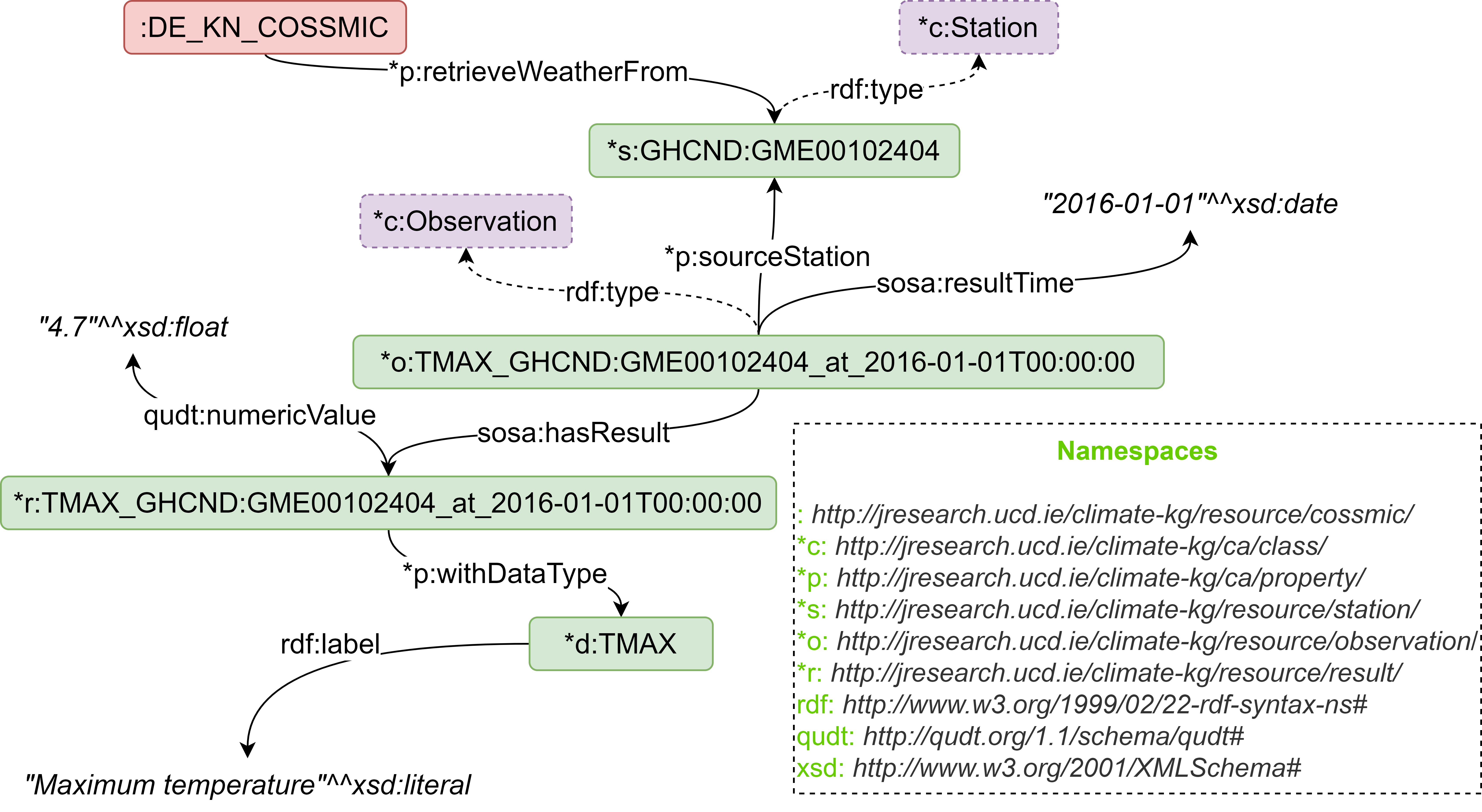}  
\caption{The red node ``DE\_KN\_COSSMIC'' is identical to the one in Fig.~\ref{fig:ontohead}; The green nodes represent modeled individuals using link-climate data, whereas the purple nodes indicate classes.}
\label{fig:hec}
\end{figure}

Using semantic methods, an energy distribution network is built in the stages of Section~\ref{sec: onto}. By specifying appropriate ontology words, it is simpler to integrate more Linked Data sources accessible on the Web. This will provide additional information that will aid in the comprehension of household energy use and production. To enable semantic integration of meteorological data into the energy network, we add a new term ``retrieveWeatherFrom'' to the CA ontology~\cite{wu2021ontology} (the ontology established in Section~\ref{sec:linkclimate} for linked climate data). As stated in Section~\ref{sec:linkclimate}, the meteorological data utilized for modeling purposes comes from our publicly available linked climate data. To illustrate the structure of the connected climate data, we provide a list of the languages used in the CA ontology and a graphical representation (Fig.~\ref{fig:hec}) of a single temperature record from a Konstanz weather station:

\begin{itemize}
    \item \textbf{*c:Station}~~a CLASS denotes a station that observes some feature of interest such as precipitation, temperature, etc.;
    \item \textbf{*c:Observation}~~a CLASS denotes an observation of some feature of interest;
    \item \textbf{*p:sourceStation}~~a PROPERTY links an observation to the station that it belongs;
    \item \textbf{*p:withDataType}~~a PROPERTY links the data to its data type;
    \item \textbf{*p:retrieveWeatherFrom}~~a PROPERTY links an individual to the individual that provides the weather information;
    \item \textbf{sosa:hasResult}~~a PROPERTY links an observation to its result.
    \item \textbf{sosa:resultTime}~~a PROPERTY links an observation to the time when the observation is generated.
    
\end{itemize}

\subsection{Conversion to Linked Data}
\label{linkeddata}
We transform the CoSSMic data from tabular to graph format, i.e. to RDF, in this step. The CoSSMic RDF data is organized using the previously developed ontology model. We write a Python script that uses RDFLib\footnote{\url{https://rdflib.readthedocs.io/en/stable/}} to convert all tabular data entries to RDF data and then put the resultant RDF in the Fuseki\footnote{\url{https://jena.apache.org/documentation/fuseki2/}} Linked Data platform. There are many compelling reasons to publish CoSSMic data as Linked Data. Any community of interest is capable of gaining access to the data through HTTP. Other data sources may advantageously integrate the CoSSMic data through Linked Data methods (\textit{e.g.} providing the data with HTTP URIs, exposing it on the Web with HTTP URIs, and establishing semantically meaningful links to the CoSSMic data) and modify the data at the most individual level, rather than downloading a large amount of CSV data. Additionally, it enables the CoSSMic data to be integrated with other information other than NOAA climate data, such as air pollution, greenhouse gas emissions, and additional meteorological data in other Linked Data platforms. Section~\ref{sec:analysishc} will show how to do queries on the Linked Data platform and provide an example of climate and CoSSMic data analysis.

\section{Linked Data platform analysis of HECP and climate data}
\label{sec:analysishc}
The purpose of this case study of linked CoSSMic HECP is to demonstrate that, on a concrete level, knowledge of fixed CoSSMic data can be enhanced through integration with the link-climate knowledge graph, and that, on a higher level of abstraction, the ontology model used to integrate cross-domain data can successfully increase the reliability of data-driven methods developed to understand decentralized energy data. Measuring the improvement on the data comprehension may take a variety of forms and is very dependent on the computational models used. We keep it basic and employ Pearson Correlation Coefficients (PCCs) to explain some additional information that may be added to the CoSSMic energy data. The SPARQL endpoint is used to acquire these two datasets (see Section~\ref{sec:query}), and the PCCs calculated for a large number of smart devices will be examined in detail.

\subsection{Semantic queries on the combination of HECP and climate data}
\label{sec:query}

The disclosed endpoint provides many methods for obtaining the linked HECP and climate data. Users may either conduct inquiries directly on the endpoint interface\footnote{\url{http://jresearch.ucd.ie/kg/dataset.html?tab=query&ds=/climate}} or incorporate them in their code as HTTP requests. The latter is supported by certain of the endpoint server's HTTP APIs and may be more efficient when doing more complex query evaluations (i.e. consuming computing resources at the client side). The queries should be prepared in the SPARQL 1.1 standard language to establish a graph pattern for obtaining the questions' answers\cite{world2013sparql,hogan2020knowledge}. Listing~\ref{lst:sparql} illustrates a query for obtaining data on the time series of solar energy produced by all CoSSMic energy device and their associated weather alignments.

\begin{lstlisting}[language=SPARQL,caption={From the public SPARQL endpoint, this example query gets a time series of CoSSMic HECP data with temperature alignments.},label=lst:sparql]
BASE <http://jresearch.ucd.ie/climate-kg/> 
PREFIX rdf: <http://www.w3.org/1999/02/22-rdf-syntax-ns#> 
PREFIX xsd: <http://www.w3.org/2001/XMLSchema#> 
PREFIX rdfs: <http://www.w3.org/2000/01/rdf-schema#> 
PREFIX seas: <https://w3id.org/seas/> 
PREFIX qudt: <http://qudt.org/1.1/schema/qudt#>
PREFIX prov: <http://www.w3.org/ns/prov#>
PREFIX sosa: <http://www.w3.org/ns/sosa/>

SELECT ?eval ?val ?maxTprt ?date
FROM <urn:x-arq:DefaultGraph> #Link-climate KG as the default graph
FROM NAMED <graph/cossmic>  #retrieve the publihshed CoSSMic graph 
WHERE
{
  ?obsv a <ca/class/Observation> ;
        <ca/property/sourceStation> <resource/station/GHCND:GME00102404> ; #code of Konstanz station
        sosa:resultTime ?date ;
        sosa:hasResult/qudt:numericValue ?maxTprt ; #maximum temperature
        sosa:hasResult/<ca/property/withDataType> <resource/datatype/TMAX> . 
  GRAPH <graph/cossmic>
  {
    <resource/cossmic/DE_KN_COSSMIC> <ca/property/retrieveWeatherFrom> <resource/station/GHCND:GME00102404>.
    <resource/cossmic/DE_KN_industrial1_pv_1> seas:evaluation ?eval. #evaluation made by CoSSMic devices
    ?eval prov:generatedAtTime ?edate;
           seas:evaluatedValue/qudt:numericalValue ?val.
  }
  
  FILTER (year(?date)=year(?edate) && month(?date)=month(?edate) && day(?date)=day(?edate))
}
#LIMIT 25 #uncomment this line to get a shorter output
\end{lstlisting}

\subsection{An example analysis of HECP against temperature}
\label{sec:hecp}
By linking HECP to climatic data, a fresh perspective on HECP can be gained. The incorporation of climate data enables a more in-depth examination of the weather's impact on energy use and production. We use PCC analysis to measure the basic linear correlation between the two sources of datasets. PCC analysis is a classic technique for selecting features based on the linearity of the variables. PCC analysis continues to be the most prevalent first step to identify the most significant features in many studies that use machine learning approaches to conduct data analysis, including the HECP~\cite{jebli2021prediction,Ciulla2019-ea}. PCC analysis is hampered, however, by its inability to comprehend nonlinear correlations and the curvature of the line. A scatter plot depicting the relationship's shape is usually advised to provide more facts on the relationships beyond the plain numerical PCC~\cite{Gogtay2017-pv}. The following Table~\ref{table:pearson} are made by selecting all the CoSSMic HECP variables that have obvious linear correlations (|PCC|$\geq$0.7) with maximum temperature (TMAX) as per day . Throughout the table, many HECP variables have positive or negative linear correlation with maximum temperature, such as solar energ generation, grid import, freezer. 

% \begin{table}[ht]
% \caption{HECP variables of PCC$\geq$0.7 with TMAX} \label{table:pearson}
% \centering
% \begin{tabular}{lr}
% \toprule
% {} &  TMAX \\
% \midrule
% DE\_KN\_industrial1\_pv\_1              &  0.78 \\
% DE\_KN\_industrial1\_pv\_2              &  0.80 \\
% DE\_KN\_industrial2\_pv                &  0.80 \\
% DE\_KN\_industrial3\_cooling\_aggregate &  0.81 \\
% DE\_KN\_industrial3\_cooling\_pumps     &  0.86 \\
% DE\_KN\_industrial3\_pv\_roof           &  0.84 \\
% DE\_KN\_residential1\_freezer          &  0.72 \\
% DE\_KN\_residential1\_heat\_pump        & -0.95 \\
% DE\_KN\_residential1\_pv               &  0.78 \\
% DE\_KN\_residential2\_circulation\_pump & -0.88 \\
% DE\_KN\_residential3\_circulation\_pump & -0.78 \\
% DE\_KN\_residential3\_freezer          &  0.75 \\
% DE\_KN\_residential4\_freezer          &  0.87 \\
% DE\_KN\_residential4\_grid\_export      &  0.82 \\
% DE\_KN\_residential4\_grid\_import      & -0.85 \\
% DE\_KN\_residential4\_heat\_pump        & -0.88 \\
% DE\_KN\_residential4\_pv               &  0.79 \\
% DE\_KN\_residential5\_refrigerator     &  0.71 \\
% DE\_KN\_residential6\_grid\_import      & -0.73 \\
% DE\_KN\_residential6\_pv               &  0.79 \\
% \bottomrule
% \end{tabular}
% \end{table}

\begin{table}[ht] \centering
\caption{HECP variables of PCC$\geq$0.7 with TMAX} \label{table:pearson}
\resizebox{\columnwidth}{!}{%
    \begin{tabular}{@{} *{21}cl @{}}
    \toprule
        & \rot{DE\_KN\_industrial1\_pv\_1} 
        & \rot{DE\_KN\_industrial1\_pv\_2} 
        & \rot{DE\_KN\_industrial2\_pv} 
        & \rot{DE\_KN\_industrial3\_cooling\_aggregate} 
        & \rot{DE\_KN\_industrial3\_cooling\_pumps} 
        & \rot{DE\_KN\_industrial3\_pv\_roof} 
        & \rot{DE\_KN\_residential1\_freezer} 
        & \rot{DE\_KN\_residential1\_heat\_pump} 
        & \rot{DE\_KN\_residential1\_pv}
        & \rot{DE\_KN\_residential2\_circulation\_pump}
        & \rot{DE\_KN\_residential3\_circulation\_pump}
        & \rot{DE\_KN\_residential3\_freezer}
        & \rot{DE\_KN\_residential4\_freezer}
        & \rot{DE\_KN\_residential4\_grid\_export}
        & \rot{DE\_KN\_residential4\_grid\_import}
        & \rot{DE\_KN\_residential4\_heat\_pump}
        & \rot{DE\_KN\_residential4\_pv}
        & \rot{DE\_KN\_residential5\_refrigerator}
        & \rot{DE\_KN\_residential6\_grid\_import}
        & \rot{DE\_KN\_residential6\_pv}\\
        \midrule
        & 0.78 &  0.80 & 0.80  & 0.81  & 0.86  & 0.84  & 0.72 & -0.95 & 0.78 & -0.88 & -0.78 & 0.75 & 0.87 & 0.82 & -0.85 & -0.88 & 0.79 & 0.71 & -0.73 & 0.79 & \textbf{TMAX}\\
        \bottomrule
    \end{tabular}
    }
\end{table}

This table only specifies the initial Pearson linear pattern for a single device. To classify the patterns of HECP variables against maximum temperature in terms of device types, we group all the CoSSMic smart devices into different device categories, calculate PCCs between each item in the same group and TMAX, and use scatter plots to verify that the calculated numerical linearity persists. To keep things simple, we'll mention three kinds of devices that are often utilized by families in CoSSMic projects: photovoltaic (PV), refrigerator/freezer, and grid import. Other device types may be examined similarly to the following sections.

\subsubsection{PV \textit{vs.} TMAX}\label{sec:pt}
\begin{figure}[ht]
\centering
\includegraphics[width=\textwidth]{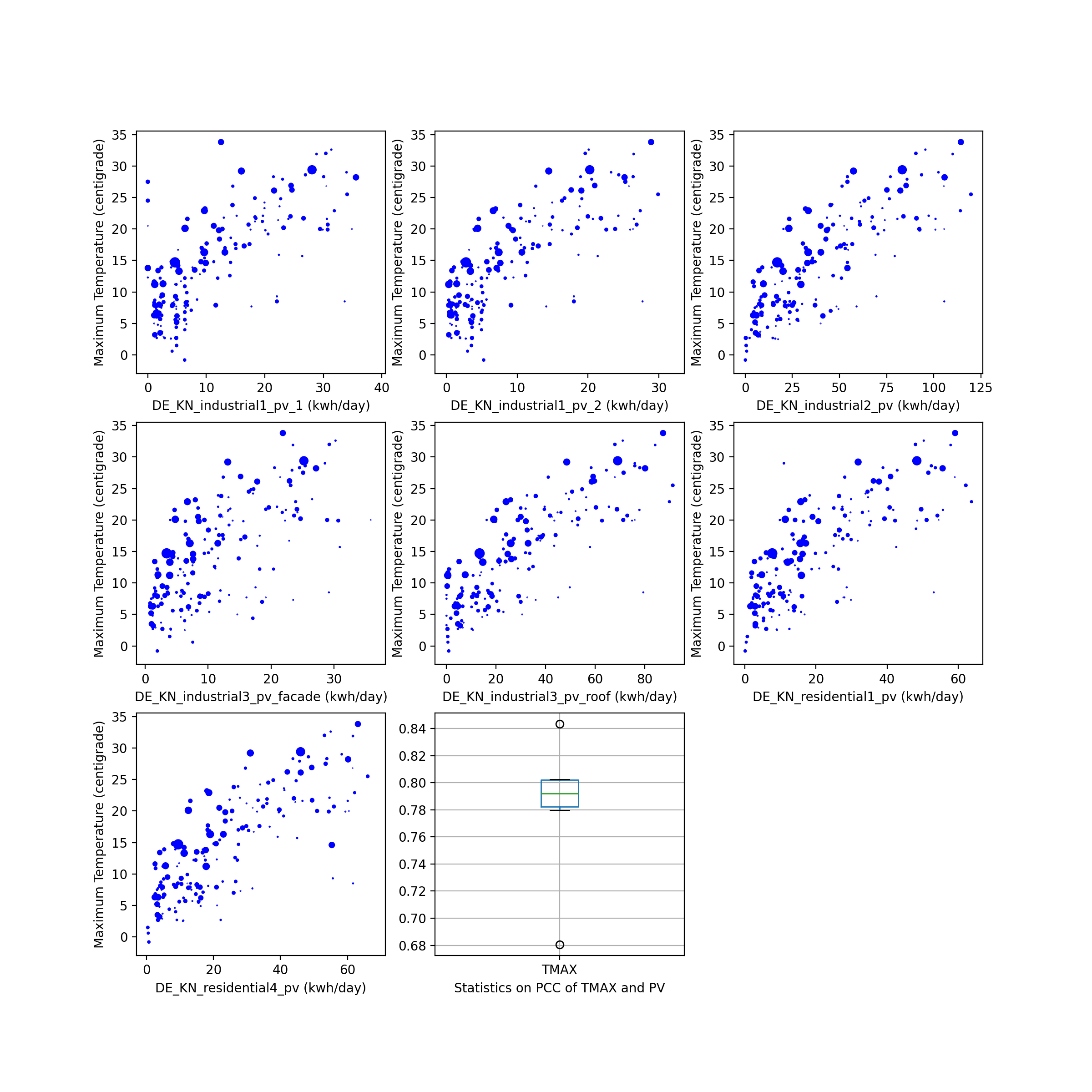} 
\caption{Daily solar energy production against daily maximum temperature, together with the quantity of precipitation in the form of scatter size}
\label{fig:pv_tmax}
\end{figure}

We commence by demonstrating how additional information may be acquired by doing a correlation study between daily maximum temperature and solar energy production by all CoSSMic PV panels. Table~\ref{tb:pvtem} contains the PCCs calculated between the solar energy generation and daily maximum temperature. It reflects a stable linear relationship between these two variables. However, to have more detailed evidence in demonstrating the linearity, we use a series of scatter plots coupled with the calculated PCC statistics box plot (Figure~\ref{fig:pv_tmax}) to solidate the linear connection between photovoltaic energy output and daily maximum temperature. As demonstrated in the scatter plots, a linear correlation pattern with a similar shape exists ( for the interaction between daily maximum temperature and solar energy production (with PCC$/approx$0.8 in the final box plot). This result is advantageous when choosing features for modeling tasks, since the maximum temperature may be one of the characteristics included to enhance modeling abilities~\cite{alskaif2020systematic, liu2020daily, manandhar2018systematic}. Additionally, the scatter color with a greater density appears in the lower range of solar energy production and maximum temperature. This suggests that PVs may produce less energy as a result of the lower temperature and more precipitation. We will not provide the complete PCC findings depending on precipitation since no CoSSMic HECP has a clear linear relationship with precipitation. Precipitation and other climatic and energy factors may have resulted in a more complicated association pattern. Additionally, one of the distinguishing features of PVs in comparison to other device types in CoSSMic is that they are the least impacted by household users, since all solar energy generated is transmitted to the central grid and subsequently distributed to the grid network for on-demand energy consumption. As a result, the linear pattern may be considered a fixed pattern in this project.
 
%  \begin{table}[ht]
%  \small
%      \centering
% \caption{PCC measurements between PV and TMAX}\label{tb:pvtem}
% \begin{tabular}{lr}
% \toprule
% PV &      TMAX \\
% \midrule
% DE\_KN\_industrial1\_pv\_1      &  0.78 \\
% DE\_KN\_industrial1\_pv\_2      &  0.80 \\
% DE\_KN\_industrial2\_pv        &  0.80 \\
% DE\_KN\_industrial3\_pv\_facade &  0.68 \\
% DE\_KN\_industrial3\_pv\_roof   &  0.84 \\
% DE\_KN\_residential1\_pv       &  0.78 \\
% DE\_KN\_residential4\_pv       &  0.79 \\
% DE\_KN\_residential6\_pv       &  0.79 \\
% \bottomrule
% \end{tabular}
%  \end{table}
 
\begin{table}[ht] \centering \footnotesize
\caption{PCC measurements between PV and TMAX}\label{tb:pvtem}
% \resizebox{\columnwidth}{!}{%
    \begin{tabular}{@{} l*{8}c @{}}
    \toprule
        & \rot[45]{DE\_KN\_industrial1\_pv\_1} 
        & \rot[45]{DE\_KN\_industrial1\_pv\_2} 
        & \rot[45]{DE\_KN\_industrial2\_pv} 
        & \rot[45]{DE\_KN\_industrial3\_pv\_facade} 
        & \rot[45]{DE\_KN\_industrial3\_pv\_roof}
        & \rot[45]{DE\_KN\_residential1\_pv} 
        & \rot[45]{DE\_KN\_residential4\_pv} & \\
        \midrule
        & 0.78 &  0.80 & 0.80  & 0.68  & 0.84  & 0.78  & 0.79 & \textbf{TMAX}\\
        \bottomrule
    \end{tabular}
    % }
\end{table}
 
\subsubsection{Refrigerator/Freezer \textit{vs.} TMAX}
\begin{figure}[ht]
\centering
\includegraphics[width=\textwidth]{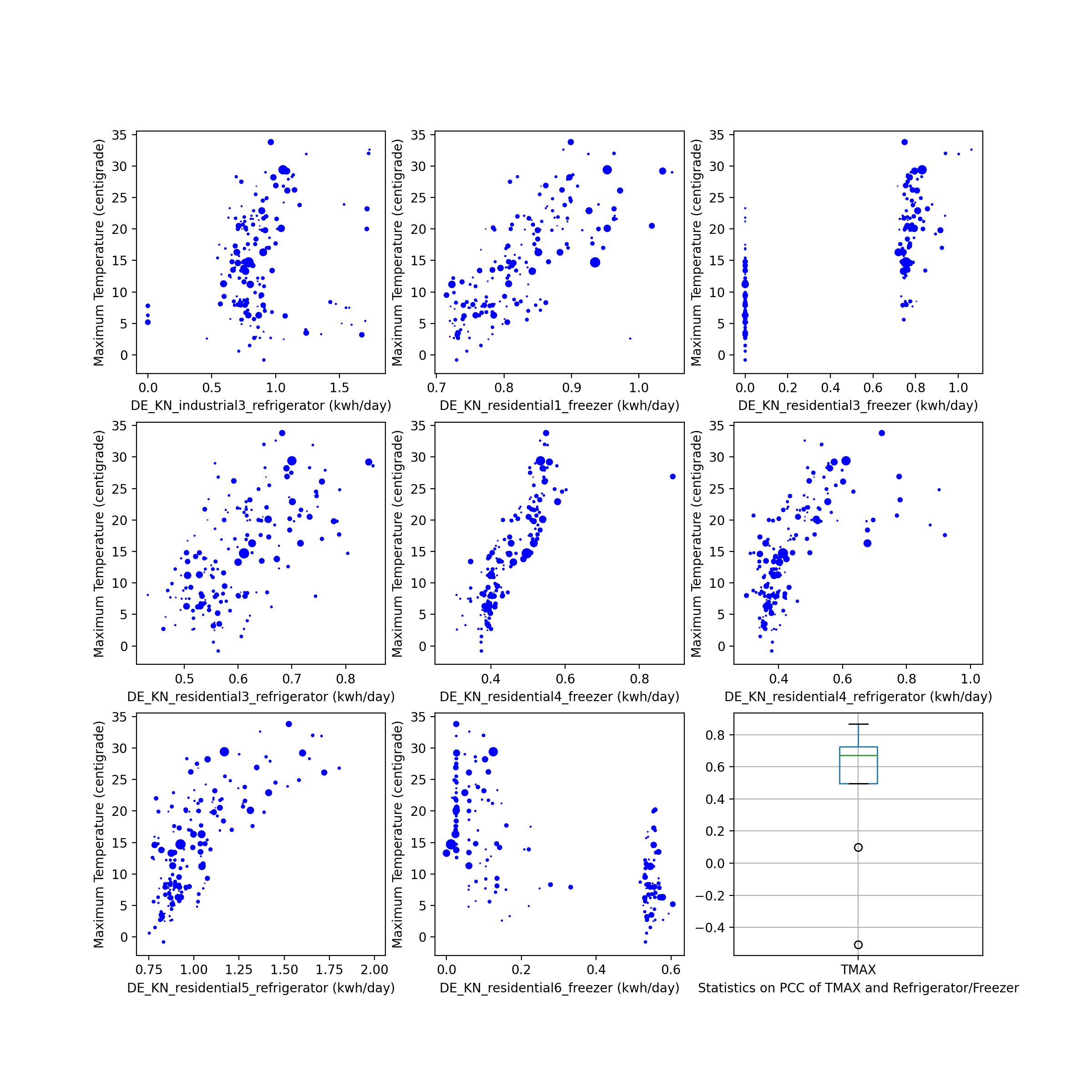} 
\caption{Daily refrigerator/freezer energy consumption against daily maximum temperature, together with the quantity of precipitation in the form of scatter size}
\label{fig:freezer_tmax}
\end{figure}

According to Table~\ref{tb:tmaxfre} of PCCs, except the ``industrial3\_refrigerator'' and ``residential6\_freezer'', a strong linear relationship exists between the refrigerator/freezer energy consumption and TMAX. However, the shape of the scatter subplot showing two vertical lines for ``residential3\_freezer'' in Figure~\ref{fig:freezer_tmax} is not consistent with the calculated PCC of 0.63 in Table~\ref{tb:tmaxfre}, which should not be simply recognized as linear correlation. The cause for non-linear relationship of these two variables may be due to that the refrigerator/freezer energy consumption \textit{vs.} TMAX is highly dependent on the households users (\textit{e.g.} food storage amount, household member number). Therefore, a universal model as illustrated in Section~\ref{sec:pt} for this pair of variables cannot be constructed using the available data.

% \begin{table}[ht]
% \small
% \centering
% \caption{PCC measurements between refrigerator/freezer and TMAX}\label{tb:tmaxfre}
% \begin{tabular}{lr}
% \toprule
% Refrigerator/freezer &      TMAX \\
% \midrule
% DE\_KN\_industrial3\_refrigerator  &  0.10 \\
% DE\_KN\_residential1\_freezer      &  0.72 \\
% DE\_KN\_residential3\_freezer      &  0.75 \\
% DE\_KN\_residential3\_refrigerator &  0.63 \\
% DE\_KN\_residential4\_freezer      &  0.87 \\
% DE\_KN\_residential4\_refrigerator &  0.64 \\
% DE\_KN\_residential5\_refrigerator &  0.71 \\
% DE\_KN\_residential6\_freezer      & -0.51\\
% \bottomrule
% \end{tabular}
% \end{table}

\begin{table}[ht] \centering \footnotesize
\caption{PCC measurements between refrigerator/freezer and TMAX}\label{tb:tmaxfre}
% \resizebox{\columnwidth}{!}{%
    \begin{tabular}{@{} l*{9}c @{}}
    \toprule
        & \rot[45]{DE\_KN\_industrial3\_refrigerator} 
        & \rot[45]{DE\_KN\_residential1\_freezer} 
        & \rot[45]{DE\_KN\_residential3\_freezer} 
        & \rot[45]{DE\_KN\_residential3\_refrigerator} 
        & \rot[45]{DE\_KN\_residential4\_freezer} 
        & \rot[45]{DE\_KN\_residential4\_refrigerator} 
        & \rot[45]{DE\_KN\_residential5\_refrigerator} 
        & \rot[45]{DE\_KN\_residential6\_freezer} & \\
        \midrule
        & 0.10 &  0.72 & 0.75  & 0.63  & 0.87  & 0.64  & 0.71 & -0.51 & \textbf{TMAX}\\
        \bottomrule
    \end{tabular}
    % }
\end{table}

\newpage
\subsubsection{Grid Import \textit{vs.} TMAX}
\begin{figure}[ht]
\centering
\includegraphics[width=\textwidth]{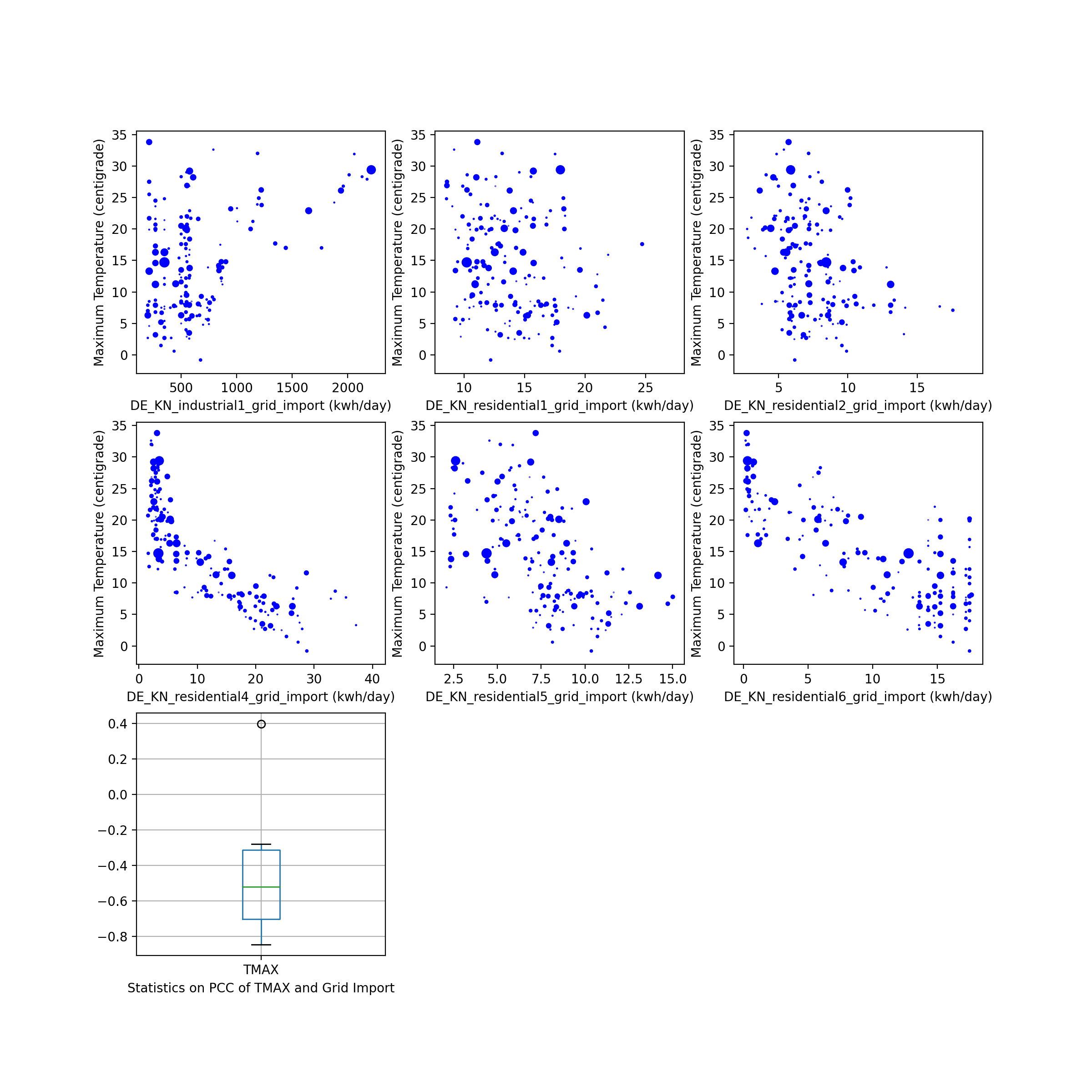} 
\caption{Daily grid import energy against daily maximum temperature, together with the quantity of precipitation in the form of scatter size}
\label{fig:grid_tmax}
\end{figure}

Similarly to the refrigerator/freezer used by households, grid import is strongly dependent on the energy consumption profile of the household, since grid import is the source energy input for all household appliances. Residential households often utilize less grid electricity at warmer temperatures, as shown in Figure~\ref{fig:grid_tmax}. This may be explained by the peculiarities of the city climate. Certain warm cities may have higher energy costs as a result of the prolonged period of high temperatures~\cite{Fung2006-qt}. There is no solid shape of the linearity (see Table 4 and Figure 8) between these two variables as the PV \textit{vs.} TMAX measured in Section~\ref{sec:pt}.

% \begin{table}[ht]
% \small
%     \centering
% \caption{PCC measurements between grid import and  TMAX}
% \begin{tabular}{lr}
% \toprule
% Grid import &      TMAX \\
% \midrule
% DE\_KN\_industrial1\_grid\_import  &  0.40 \\
% DE\_KN\_residential1\_grid\_import & -0.28 \\
% DE\_KN\_residential2\_grid\_import & -0.42 \\
% DE\_KN\_residential4\_grid\_import & -0.85 \\
% DE\_KN\_residential5\_grid\_import & -0.62 \\
% DE\_KN\_residential6\_grid\_import & -0.73 \\
% \bottomrule
% \end{tabular}
% \end{table}

\begin{table}[ht] \centering \footnotesize
\caption{PCC measurements between grid import and  TMAX}
% \resizebox{\columnwidth}{!}{%
    \begin{tabular}{@{} l*{7}c @{}}
    \toprule
        & \rot[45]{DE\_KN\_industrial1\_grid\_import} 
        & \rot[45]{DE\_KN\_residential1\_grid\_import} 
        & \rot[45]{DE\_KN\_residential2\_grid\_import} 
        & \rot[45]{DE\_KN\_residential4\_grid\_import} 
        & \rot[45]{DE\_KN\_residential5\_grid\_import} 
        & \rot[45]{DE\_KN\_residential6\_grid\_import} \\
        \midrule
        & 0.40 &  -0.28 & -0.42  & -0.85  & -0.62  & -0.73  & \textbf{TMAX}\\
        \bottomrule
    \end{tabular}
    % }
\end{table}

\section{Discussion}
This section discusses the value of semantic approaches in terms of enhancing a local energy dataset. The technique used in this study is sample research on the interplay of energy and climate data. However, because of the ontology model's higher-level abstraction, this method may be extended to account for the various interests of multiple groups of people, including energy companies, system operators, dealers, and policy makers. Whatever purpose this strategy is being utilized for, there are certain general advantages and disadvantages concerning the following facets to be aware of.

\subsection{Adding new datasets}
\label{para:addnew}
A new dataset must be defined and linked to an ontology in order to be included in a SPARQL endpoint as part of the Web of data (or Linked Data), in a manner similar to the process described in this article. However, since the schema (or ontology) for RDF triples is not standardized, users interested in using the data provided by other Linked Data sources must also understand the ontology of the data in order to construct graph patterns using the SPARQL language to identify the desired data. Normally Linked Data should include itemized documentations to aid users in either updating data or making SPARQL queries.

\subsection{Interlinking diverse data}
\label{para:interlink}
In many scenarios, such as cross-domain analysis, heterogeneous data sources must be merged into a single dataset. At the ontology level, semantically linked heterogeneous data may be merged effectively with other specialized Linked Data fields. However, certain crucial disadvantages will endure for years. First, it requires a sufficient amount of domain knowledge in the schemas of all relevant datasets, as well as a number of acceptable ontology vocabularies for semantic enrichment~\cite{patricio2020web,musyaffa2020iota}. Second, it will significantly increase the complexity of the data as the ontologies created by various parties are challenged to be united in compliance with some common protocol. The vocabulary selected for the same associated concept might differ substantially. Thus, ontology creators should take full benefit of the existing ontologies either by reusing them or linking them to their vocabularies.

\subsection{Increasing data collection to better understand smart energy} 
We show in this study how one additional data source (NOAA meteorological variables) may be used to enhance our knowledge of a smart energy system in Konstanz. Numerous additional data pieces, such as solar radiation data\footnote{\url{https://www.dwd.de/EN/ourservices/solarenergy/maps_globalradiation_average.html}}, other meteorological data (e.g. meteoblue\footnote{\url{https://www.meteoblue.com/en/weather/archive/export/konstanz_germany_2885679}}), and points of interest (e.g. OpenStreetMap\footnote{\url{https://www.openstreetmap.org/}}), may also be absorbed to this purpose. Assimilation of these datasets should adhere to the aforementioned criteria in Section~\ref{para:addnew} and Section~\ref{para:interlink}.

\subsection{Federated queries across SPARQL endpoints}\label{sec:fedsparql}
The data that is published as Linked Data does not have to be contained in a single SPARQL endpoint. SPARQL endpoints implemented in standard or extended SPARQL allow federated queries with other endpoints, \textit{i.e.}, data spread over many distant endpoints may be utilized directly as sources and retrieved in any endpoint through SPARQL queries\cite{grall2020collaborative}. Linked Open Data\footnote{\url{https://lod-cloud.net/}} has collected a collection of open Linked data sources from which users may obtain data useful for energy data processing.

\section{Conclusion and Future Work}
\label{sec:Conc}
We provide a process for converting a local decentralized household energy consumption and production dataset (CoSSMic) to a Linked dataset and republishing it over the web in this article. This study details the procedures necessary to accomplish this transformation, and as a consequence, the scope of the local dataset is extended, and wider perspectives of meteorological information are added to the dataset through semantic approaches. The final study of this dataset using temperature and precipitation data at the device level shows how different sources of meteorological data may be beneficial for data-driven modeling jobs. In the future, we intend to work on additional semantic methods such as GeoSPARQL and temporal RDF, as well as a broader variety of energy-related datasets such as remote sensing data, energy markets, and policy data, in order to develop a robust interoperable ontology model capable of serving a nearly complete knowledge-sharing energy data ecosystem spanning multiple domains, thereby improving understanding of the decentralized energy distribution mechanism.

\section*{Acknowledgments}
This research was partially funded by the EU H2020 research and innovation programme under the Marie Skłodowska-Curie Grant Agreement No.\ 713567 at the ADAPT SFI Research Centre at Trinity College Dublin. The ADAPT Centre for Digital Content Technology is partially supported by the SFI Research Centres Programme (Grant 13/RC/2106\_P2) and is co-funded under the European Regional Development Fund.

%\bibliography{}

\begin{thebibliography}{10}
\expandafter\ifx\csname url\endcsname\relax
  \def\url#1{\texttt{#1}}\fi
\expandafter\ifx\csname urlprefix\endcsname\relax\def\urlprefix{URL }\fi
\expandafter\ifx\csname href\endcsname\relax
  \def\href#1#2{#2} \def\path#1{#1}\fi

\bibitem{noauthor_undated-ri}
{EUR-Lex} - {52021PC0558} - {EN} - {EUR-Lex},
  \url{https://eur-lex.europa.eu/legal-content/EN/TXT/?uri=CELEX:52021PC0558},
  accessed: 2021-11-21.

\bibitem{jain2022deep}
M.~Jain, T.~AlSkaif, S.~Dev, Are deep learning models more effective against
  traditional models for load demand forecasting?, 2022.

\bibitem{wu2021organizing}
J.~Wu, F.~Orlandi, Y.~H. Lee, D.~O’Sullivan, S.~Dev, Organizing decentralized
  energy data using semantic approach, in: 2021 Photonics \& Electromagnetics
  Research Symposium (PIERS), IEEE, 2021, pp. 2213--2216.

\bibitem{van2020integrated}
G.~van Leeuwen, T.~AlSkaif, M.~Gibescu, W.~van Sark, An integrated
  blockchain-based energy management platform with bilateral trading for
  microgrid communities, Applied Energy 263 (2020) 114613.

\bibitem{Xu2020-xg}
Y.~Xu, C.~Yan, H.~Liu, J.~Wang, Z.~Yang, Y.~Jiang, Smart energy systems: A
  critical review on design and operation optimization, Sustainable Cities and
  Society 62 (2020) 102369.

\bibitem{jain2021validating}
M.~Jain, T.~Alskaif, S.~Dev, Validating clustering frameworks for electric load
  demand profiles, IEEE Transactions on Industrial Informatics (2021).

\bibitem{jain2020clustering}
M.~Jain, T.~AlSkaif, S.~Dev, A clustering framework for residential electric
  demand profiles, in: 2020 International Conference on Smart Energy Systems
  and Technologies (SEST), IEEE, 2020, pp. 1--6.

\bibitem{Teixeira2020-pb}
B.~Teixeira, G.~Santos, T.~Pinto, Z.~Vale, J.~M. Corchado, Application ontology
  for {Multi-Agent} and {Web-Services'} {Co-Simulation} in power and energy
  systems, IEEE Access 8 (2020) 81129--81141.

\bibitem{alskaif2020systematic}
T.~AlSkaif, S.~Dev, L.~Visser, M.~Hossari, W.~van Sark, A systematic analysis
  of meteorological variables for pv output power estimation, Renewable Energy
  153 (2020) 12--22.

\bibitem{Mussard2017-xb}
M.~Mussard, Solar energy under cold climatic conditions: A review, Renewable
  Sustainable Energy Rev. 74 (2017) 733--745.

\bibitem{dev2019estimating}
S.~Dev, F.~M. Savoy, Y.~H. Lee, S.~Winkler, Estimating solar irradiance using
  sky imagers, Atmospheric Measurement Techniques 12~(10) (2019) 5417--5429.

\bibitem{dev2016estimation}
S.~Dev, F.~M. Savoy, Y.~H. Lee, S.~Winkler, Estimation of solar irradiance
  using ground-based whole sky imagers, 2016, pp. 7236--7239.

\bibitem{orlandi2019interlinking}
F.~Orlandi, A.~Meehan, M.~Hossari, S.~Dev, D.~O'Sullivan, T.~AlSkaif,
  Interlinking heterogeneous data for smart energy systems, in: 2019
  International Conference on Smart Energy Systems and Technologies (SEST),
  IEEE, 2019, pp. 1--6.

\bibitem{wu2021interoperable}
J.~Wu, H.~Chen, F.~Orlandi, Y.~H. Lee, D.~O’Sullivan, S.~Dev, An
  interoperable open data portal for climate analysis, in: 2021 IEEE USNC-URSI
  Radio Science Meeting (Joint with AP-S Symposium), IEEE, 2021, pp. 104--105.

\bibitem{wu2021detecting}
J.~Wu, F.~Orlandi, D.~O’Sullivan, S.~Dev, Detecting rainfall events
  leveraging climate knowledge graphs, in: 2021 Photonics \& Electromagnetics
  Research Symposium (PIERS), IEEE, 2021, pp. 2336--2341.

\bibitem{wu2022rdfstr}
J.~Wu, F.~Orlandi, D.~O'Sullivan, S.~Dev, A Workflow To Convert Live
  Atmospheric Sensor Data Into Linked Data, 2022.

\bibitem{wu2021ontologyd}
J.~Wu, F.~Orlandi, T.~AlSkaif, D.~O'Sullivan, S.~Dev, Ontology modeling for
  decentralized household energy systems, in: 2021 International Conference on
  Smart Energy Systems and Technologies (SEST), IEEE, 2021, pp. 1--6.

\bibitem{ahmad2020review}
T.~Ahmad, H.~Zhang, B.~Yan, A review on renewable energy and electricity
  requirement forecasting models for smart grid and buildings, Sustainable
  Cities and Society 55 (2020) 102052.

\bibitem{wu2022boosting}
J.~Wu, F.~Orlandi, D.~O'Sullivan, E.~Pisoni, S.~Dev, Boosting climate analysis
  with semantically uplifted knowledge graphs, IEEE Journal of Selected Topics
  in Applied Earth Observations and Remote Sensing (2022).

\bibitem{wu2021automated}
J.~Wu, H.~Chen, F.~Orlandi, Y.~H. Lee, D.~O’Sullivan, S.~Dev, Automated
  climate analyses using knowledge graph, in: 2021 IEEE USNC-URSI Radio Science
  Meeting (Joint with AP-S Symposium), IEEE, 2021, pp. 106--107.

\bibitem{wu2022augment}
J.~Wu, F.~Orlandi, M.~S. Pathan, D.~O'Sullivan, S.~Dev, Augmenting Weather
  Sensor Data With Remote Knowledge Graphs, 2022.

\bibitem{Salatino2018-xm}
A.~A. Salatino, T.~Thanapalasingam, A.~Mannocci, F.~Osborne, E.~Motta, The
  computer science ontology: A {Large-Scale} taxonomy of research areas, in:
  The Semantic Web -- {ISWC} 2018, Springer International Publishing, 2018, pp.
  187--205.

\bibitem{wu2021uplifting}
J.~Wu, F.~Orlandi, I.~Gollini, E.~Pisoni, S.~Dev, Uplifting air quality data
  using knowledge graph, in: 2021 Photonics \& Electromagnetics Research
  Symposium (PIERS), IEEE, 2021, pp. 2347--2350.

\bibitem{wu2022linkclimate}
J.~Wu, F.~Orlandi, D.~O'Sullivan, S.~Dev, Link climate: An interoperable
  knowledge graph platform for climate data, Computers and Geosciences (2022).

\bibitem{Hooda2020-zq}
D.~Hooda, R.~Rani, Ontology driven human activity recognition in heterogeneous
  sensor measurements, J. Ambient Intell. Humaniz. Comput. 11~(12) (2020)
  5947--5960.

\bibitem{wu2022vkg}
J.~Wu, F.~Orlandi, D.~O'Sullivan, S.~Dev, Publishing Climate Data As Linked
  Data Via Virtual Knowledge Graphs, 2022.

\bibitem{Horrocks2004-jw}
I.~Horrocks, P.~F. Patel-Schneider, H.~Boley, S.~Tabet, B.~Grosof, M.~Dean,
  {Others}, {SWRL}: A semantic web rule language combining {OWL} and {RuleML},
  W3C Member submission 21~(79) (2004) 1--31.

\bibitem{manandhar2019data}
S.~Manandhar, S.~Dev, Y.~H. Lee, Y.~S. Meng, S.~Winkler, A data-driven approach
  for accurate rainfall prediction, IEEE Transactions on Geoscience and Remote
  Sensing 57~(11) (2019) 9323--9331.

\bibitem{manandhar2018data}
S.~Manandhar, S.~Dev, Y.~H. Lee, Y.~S. Meng, S.~Winkler, A data-driven approach
  to detect precipitation from meteorological sensor data, 2018, pp.
  3872--3875.

\bibitem{bizer2011linked}
C.~Bizer, T.~Heath, T.~Berners-Lee, Linked data: The story so far, in: Semantic
  services, interoperability and web applications: emerging concepts, IGI
  Global, 2011, pp. 205--227.

\bibitem{Abid2018-mh}
T.~Abid, M.~R. Laouar, Using semantic web and linked data for integrating and
  publishing data in smart cities, in: Proceedings of the 7th International
  Conference on Software Engineering and New Technologies, no. Article 32 in
  ICSENT 2018, Association for Computing Machinery, New York, NY, USA, 2018,
  pp. 1--4.

\bibitem{An2020-kc}
J.~An, S.~Kumar, J.~Lee, S.~Jeong, J.~Song, Synapse : Towards linked data for
  smart cities using a semantic annotation framework, in: 2020 {IEEE} 6th World
  Forum on Internet of Things ({WF-IoT}), 2020, pp. 1--6.

\bibitem{Cimmino2020-oj}
A.~Cimmino, N.~Andreadou, A.~Fern{\'a}ndez-Izquierdo, C.~Patsonakis, A.~C.
  Tsolakis, A.~Lucas, D.~Ioannidis, E.~Kotsakis, D.~Tzovaras,
  R.~Garc{\'\i}a-Castro, Semantic interoperability for {DR} schemes employing
  the {SGAM} framework, in: 2020 International Conference on Smart Energy
  Systems and Technologies ({SEST}), ieeexplore.ieee.org, 2020, pp. 1--6.

\bibitem{Fernandez-Izquierdo2020-cf}
A.~Fern{\'a}ndez-Izquierdo, A.~Cimmino, C.~Patsonakis, A.~C. Tsolakis,
  R.~Garc{\'\i}a-Castro, D.~Ioannidis, D.~Tzovaras, {OpenADR} ontology:
  Semantic enrichment of demand response strategies in smart grids, in: 2020
  International Conference on Smart Energy Systems and Technologies ({SEST}),
  2020, pp. 1--6.

\bibitem{Baken2020-ye}
N.~Baken, Linked data for smart homes: Comparing {RDF} and labeled property
  graphs, in: {LDAC2020---8th} Linked Data in Architecture and Construction
  Workshop, linkedbuildingdata.net, 2020, pp. 23--36.

\bibitem{Chun2020-su}
S.~Chun, J.~Jung, X.~Jin, S.~Seo, K.-H. Lee, Designing an integrated knowledge
  graph for smart energy services, J. Supercomput. 76~(10) (2020) 8058--8085.

\bibitem{Wagner2010-xp}
A.~Wagner, S.~Speiser, O.~Raabe, A.~Harth, Linked data for a privacy-aware
  smart grid, INFORMATIK 2010. Service (2010).

\bibitem{noauthor_undated-gh}
L.~Daniele, M.~Solanki, F.~den Hartog, J.~Roes, Interoperability for smart
  appliances in the {IoT} world (2016) 21--29.

\bibitem{Lefrancois2017-ag}
M.~Lefran{\c c}ois, Planned {ETSI} {SAREF} extensions based on the {W3C\&OGC}
  {SOSA/SSN-compatible} {SEAS} ontology paaerns, in: Workshop on Semantic
  Interoperability and Standardization in the {IoT}, {SIS-IoT}, 2017, p. 11p.

\bibitem{wu2021ontology}
J.~Wu, F.~Orlandi, D.~O’Sullivan, S.~Dev, An ontology model for climatic data
  analysis, in: Proc. IEEE International Geoscience and Remote Sensing
  Symposium (IGARSS), 2021.

\bibitem{Janowicz2019-hn}
K.~Janowicz, A.~Haller, S.~J.~D. Cox, D.~Le~Phuoc, M.~Lefran{\c c}ois, {SOSA}:
  A lightweight ontology for sensors, observations, samples, and actuators,
  Journal of Web Semantics 56 (2019) 1--10.

\bibitem{amato2017simulation}
A.~Amato, R.~Aversa, B.~Di~Martino, M.~Scialdone, S.~Venticinque, A simulation
  approach for the optimization of solar powered smart migro-grids, in:
  Conference on Complex, Intelligent, and Software Intensive Systems, Springer,
  2017, pp. 844--853.

\bibitem{Manola2004-jx}
F.~Manola, E.~Miller, B.~McBride, {Others}, {RDF} primer, W3C recommendation
  10~(1-107) (2004) 6.

\bibitem{barbosa2021use}
A.~Barbosa, I.~I. Bittencourt, S.~W.~M. Siqueira, R.~de~Amorim~Silva,
  I.~Calado, The use of software tools in linked data publication and
  consumption: A systematic literature review, Research Anthology on Digital
  Transformation, Organizational Change, and the Impact of Remote Work (2021)
  1868--1888.

\bibitem{Gomes2016-ju}
L.~Gomes, M.~Lefran{\c c}ois, P.~Faria, Z.~Vale, Publishing real-time microgrid
  consumption data on the web of linked data, in: 2016 Clemson University Power
  Systems Conference ({PSC}), 2016, pp. 1--8.

\bibitem{Wicaksono2021-fg}
{Wicaksono}, {Boroukhian}, {Bashyal}, A {Demand-Response} system for
  sustainable manufacturing using linked data and machine learning, Dynamics in
  Logistics (2021).

\bibitem{8706177}
F.~{Orlandi}, J.~{Debattista}, I.~A. {Hassan}, C.~{Conran}, M.~{Latifi},
  M.~{Nicholson}, F.~A. {Salim}, D.~{Turner}, O.~{Conlan}, D.~{O'sullivan},
  J.~{Tang}, Leveraging knowledge graphs of movies and their content for
  web-scale analysis, in: 2018 14th International Conference on Signal-Image
  Technology Internet-Based Systems (SITIS), 2018, pp. 609--616.

\bibitem{seasd22}
M.~Lefran\c{c}ois, J.~Kalaoja, T.~Ghariani, A.~Zimmermann, {SEAS Knowledge
  Model}, Deliverable 2.2, ITEA2 12004 Smart Energy Aware Systems, 76 p.
  (2016).

\bibitem{beckett2014rdf}
D.~Beckett, T.~Berners-Lee, E.~Prud’hommeaux, G.~Carothers, Rdf 1.1 turtle,
  World Wide Web Consortium (2014) 18--31.

\bibitem{world2013sparql}
W.~W.~W. Consortium, et~al., {SPARQL} 1.1 overview (2013).

\bibitem{hogan2020knowledge}
A.~Hogan, E.~Blomqvist, M.~Cochez, C.~d'Amato, G.~de~Melo, C.~Gutierrez,
  J.~E.~L. Gayo, S.~Kirrane, S.~Neumaier, A.~Polleres, et~al., Knowledge
  graphs, arXiv preprint arXiv:2003.02320 (2020).

\bibitem{jebli2021prediction}
I.~Jebli, F.-Z. Belouadha, M.~I. Kabbaj, A.~Tilioua, Prediction of solar energy
  guided by pearson correlation using machine learning, Energy 224 (2021)
  120109.

\bibitem{Ciulla2019-ea}
G.~Ciulla, A.~D'Amico, Building energy performance forecasting: A multiple
  linear regression approach, Appl. Energy 253 (2019) 113500.

\bibitem{Gogtay2017-pv}
N.~J. Gogtay, U.~M. Thatte, Principles of correlation analysis, J. Assoc.
  Physicians India 65~(3) (2017) 78--81.

\bibitem{liu2020daily}
Y.~Liu, Y.~Mu, K.~Chen, Y.~Li, J.~Guo, Daily activity feature selection in
  smart homes based on pearson correlation coefficient, Neural Processing
  Letters (2020) 1--17.

\bibitem{manandhar2018systematic}
S.~Manandhar, S.~Dev, Y.~H. Lee, S.~Winkler, Y.~S. Meng, Systematic study of
  weather variables for rainfall detection, in: Proc. IEEE International
  Geoscience and Remote Sensing Symposium (IGARSS), IEEE, 2018, pp. 3027--3030.

\bibitem{Fung2006-qt}
W.~Y. Fung, K.~S. Lam, W.~T. Hung, S.~W. Pang, Y.~L. Lee, Impact of urban
  temperature on energy consumption of hong kong, Energy 31~(14) (2006)
  2623--2637.

\bibitem{patricio2020web}
H.~S. Patr{\'\i}cio, M.~I. Cordeiro, P.~N. Ramos, From the web of bibliographic
  data to the web of bibliographic meaning: structuring, interlinking and
  validating ontologies on the semantic web, International Journal of Metadata,
  Semantics and Ontologies 14~(2) (2020) 124--134.

\bibitem{musyaffa2020iota}
F.~A. Musyaffa, M.-E. Vidal, F.~Orlandi, J.~Lehmann, H.~Jabeen, Iota:
  Interlinking of heterogeneous multilingual open fiscal data, Expert Systems
  with Applications 147 (2020) 113135.

\bibitem{grall2020collaborative}
A.~Grall, H.~Skaf-Molli, P.~Molli, M.~Perrin, Collaborative {SPARQL} query
  processing for decentralized semantic data, in: International Conference on
  Database and Expert Systems Applications, Springer, 2020, pp. 320--335.

\end{thebibliography}
% 1. Automatic Recognition of Student Engagement using Deep Learning and Facial Expression Omid Mohamad Nezami, Len Hame, Deborah Richard,
% Stephen Wan, d Cecile Pari

\end{document}